\documentclass{article}
\usepackage{pgfplots}
\pgfplotsset{compat=1.18}
\usepackage{arxiv}
\usepackage{graphicx}   % For including images
\usepackage{hyperref}   % For PDF metadata and hyperlinks
\usepackage{ifpdf}      % To handle both LaTeX and PDFLaTeX
\usepackage{caption}    % Better figure captions
\usepackage{float}      % To control figure placement
\usepackage{amsmath}    % For cases environment and math alignment
\usepackage{booktabs}

\usepackage{pgfplots}
\usepackage{csvsimple}
\usepackage{subcaption}

\ifpdf
    \pdfoutput=1
\fi

\title{Preliminary Study on Space Utilization and Emergent Behaviors of Group vs. Single Pedestrians in Real-World Trajectories}

\author{
  Sanjjamts Amartaivan, Hiroshi Morita \\
  Department of Information and Physical Sciences\\
  Graduate School of Information Science and Technology\\
  The University of Osaka, Osaka, Japan\\
  \texttt{\{s.amaru, morita\}@ist.osaka-u.ac.jp} 
  \\
  }

\date{\today}

\begin{document}
\maketitle

\begin{abstract}
This study presents an initial framework for distinguishing group and single pedestrians based on real-world trajectory data, with the aim of analyzing their differences in space utilization and emergent behavioral patterns. By segmenting pedestrian trajectories into fixed time bins and applying a Transformer-based pair classification model, we identify cohesive groups (flocks) and isolate single pedestrians over a structured sequence-based filtering process.

To prepare for deeper analysis, we establish a comprehensive metric framework incorporating both spatial and behavioral dimensions. Spatial utilization metrics include convex hull area, smallest enclosing circle (SEC) radius, and heatmap-based spatial densities to characterize how different pedestrian types occupy and interact with space. Behavioral metrics such as velocity change ($\Delta v$), motion angle deviation ($\Delta \theta$), clearance radius, and trajectory straightness are designed to capture local adaptations and responses during interactions. Furthermore, we introduce a typology of encounter types—single-to-single (S–S), single-to-group (S–G), and group-to-group (G–G)—to categorize and later quantify different interaction scenarios.

Although this version focuses primarily on the classification pipeline and dataset structuring, it establishes the groundwork for scalable analysis across different sequence lengths (e.g., 60, 100, and 200 frames). Future versions will incorporate complete quantitative analysis of the proposed metrics and their implications for pedestrian simulation and space design validation in crowd dynamics research.
\end{abstract}

\section{Introduction}

Understanding how pedestrians utilize space in shared environments is a key concern in both crowd dynamics research and urban design. Pedestrian flow is not only dictated by individual movement intentions but also by interactions with surrounding agents and environmental constraints. Among these, group behaviors add layers of complexity, especially when pedestrians move in cohesive formations. The way groups occupy space and respond to surrounding conditions can significantly influence the overall efficiency and safety of public facilities such as corridors, transport hubs, and event venues.

Efficient space utilization is critical for both pedestrian experience and architectural design validation. Poorly optimized layouts can lead to congestion, navigation delays, or unsafe conditions during peak hours or emergency evacuations. As a result, designers and simulation modelers must carefully account for how pedestrians — particularly those in groups — move, interact, and adapt within spatial constraints. Traditional design guidelines often assume independent pedestrian movement, potentially overlooking the emergent patterns caused by coordinated group dynamics. Thus, validating design elements against realistic group behavior is essential for creating inclusive and effective public infrastructures.

Collective motion in trajectory data can be categorized into different formats, including \textbf{flocks}, \textbf{convoys}, and \textbf{swarms}~\cite{wang2020big}. A \textbf{flock} is a set of agents moving together within a limited spatial region over a specific time interval. A \textbf{convoy} extends this definition by maintaining the same group structure over longer periods, making it more stable in dynamic environments. A \textbf{swarm} represents a more loosely connected group, where individuals exhibit similar movement patterns but do not necessarily maintain fixed spatial relationships. In pedestrian contexts, flocks may correspond to families, friends, or crew members walking together with implicit coordination and mutual awareness.

In this study, we investigate how group and single pedestrians utilize space differently in real-world environments and explore the behaviors that emerge from their interactions. By classifying pedestrian trajectories into group or single categories over temporal windows, we analyze their movement footprints, interactive dynamics, and local adaptations in shared space. Visual inspection of trajectory data reveals that group pedestrians often attempt to maintain their flock formations, even in constrained or crowded conditions. This flock-preserving behavior contributes to spatial dominance, where group agents occupy key pathways or choke points, impeding the movement of single pedestrians.

In addition to analyzing spatial utilization patterns, we extract emergent behavioral strategies such as detouring, yielding, aggressive passing, and flock maintenance. These behaviors provide insight into how pedestrians respond to complex spatial and social configurations in real time. By combining spatial metrics with behavior-level observations, we aim to deliver a comprehensive understanding of pedestrian strategies in mixed environments.

Our findings contribute to the advancement of behaviorally grounded pedestrian simulation models, enabling better prediction of group dynamics in dense scenarios. Moreover, the results offer empirical evidence for validating architectural layouts and pedestrian flow designs, ultimately supporting the development of more robust and inclusive public spaces.

\section{Literature Review}

Research in pedestrian dynamics has evolved significantly with the availability of trajectory-level datasets, allowing detailed analysis of both individual and group behaviors in various environments. A growing body of work emphasizes the need to account for group behavior when modeling and designing public spaces, as group dynamics can substantially affect flow efficiency and space occupation.

\textbf{Group pedestrian behavior} has been widely studied in the context of simulation and observation. Moussaïd et al.~\cite{moussaid2010walking} highlighted the social interactions and self-organizing principles that drive group formation and movement, showing that group pedestrians typically walk side-by-side and adjust their velocity to maintain cohesion. Zanlungo et al.~\cite{zanlungo2012social} proposed models that capture social forces and proxemic behavior to explain how groups maintain inter-member distances while navigating shared spaces.

From a simulation standpoint, studies have incorporated group dynamics into agent-based models to more realistically capture space utilization and flow patterns. For example, Schultz and Fricke~\cite{schultz2010group} introduced group-oriented behavior rules into microscopic simulations, demonstrating how group cohesion alters fundamental diagrams and flow-density relations. These models support design validation by allowing what-if analyses of spatial arrangements under group-dominated flows.

In the domain of \textbf{collective motion mining}, trajectory pattern analysis has led to the classification of movement behaviors into flocks, swarms, and convoys. Wang et al.~\cite{wang2020big} provided formal definitions of these group movement types using spatio-temporal constraints, enabling the mining of cohesive movement patterns from large-scale trajectory datasets. These categorizations are crucial in understanding not only how groups move, but also how persistently they maintain spatial proximity over time.

In terms of \textbf{space utilization and environmental design}, several studies have focused on the implications of pedestrian grouping. Hoogendoorn and Daamen~\cite{hoogendoorn2005pedestrian} investigated the influence of group behavior on bottleneck flow and evacuation efficiency, highlighting how tightly-knit groups can increase exit time in constrained layouts. In architectural validation, Karamouzas et al.~\cite{karamouzas2014universal} stressed the importance of integrating empirical data into simulations to assess the realism and robustness of pedestrian facility designs.

Despite these advances, relatively few studies focus explicitly on the \textbf{comparative analysis of group and single pedestrian space usage} using empirical data. Our study aims to address this gap by offering visual and quantitative insights from real-world trajectories, identifying patterns where group pedestrians dominate shared spaces, and validating whether existing simulation assumptions sufficiently account for such behavior.

\section{Methodology}

\subsection{Dataset}

We use a real-world pedestrian group-identified trajectory dataset \cite{atc_track} from an indoor environment recorded at the Asian Trade Center in Osaka, Japan.

Each row in the CSV file corresponds to a single tracked individual at a specific time instant and includes the following fields: \textit{time [ms] (Unix time + milliseconds/1000)}, \textit{person ID}, \textit{position x [mm]}, \textit{position y [mm]}, \textit{velocity [mm/s]}, \textit{angle of motion [rad]}, and \textit{facing angle [rad]}.

For this study, we selected trajectory data within the time interval 12:00–13:00,from a single day—May 8, 2013 (Wednesday)—within a 7-day observation period that includes manually annotated group information, to conduct proposed analysis. Additionally, data from three Sundays (February 17, March 24, and May 5, 2013), which contain labeled pedestrian pairs, were used to train a pair classification model based on sequential deep learning architectures, specifically a Transformer model. Our focus was on the eastern corridor region, an area characterized by dense and compact pedestrian flow. This corridor includes several points of interest, such as exits, retail spaces with varying levels of attraction, and structural features like pillars that influence movement patterns.

\begin{figure}[H]
\centering
\includegraphics[width=0.8\linewidth]{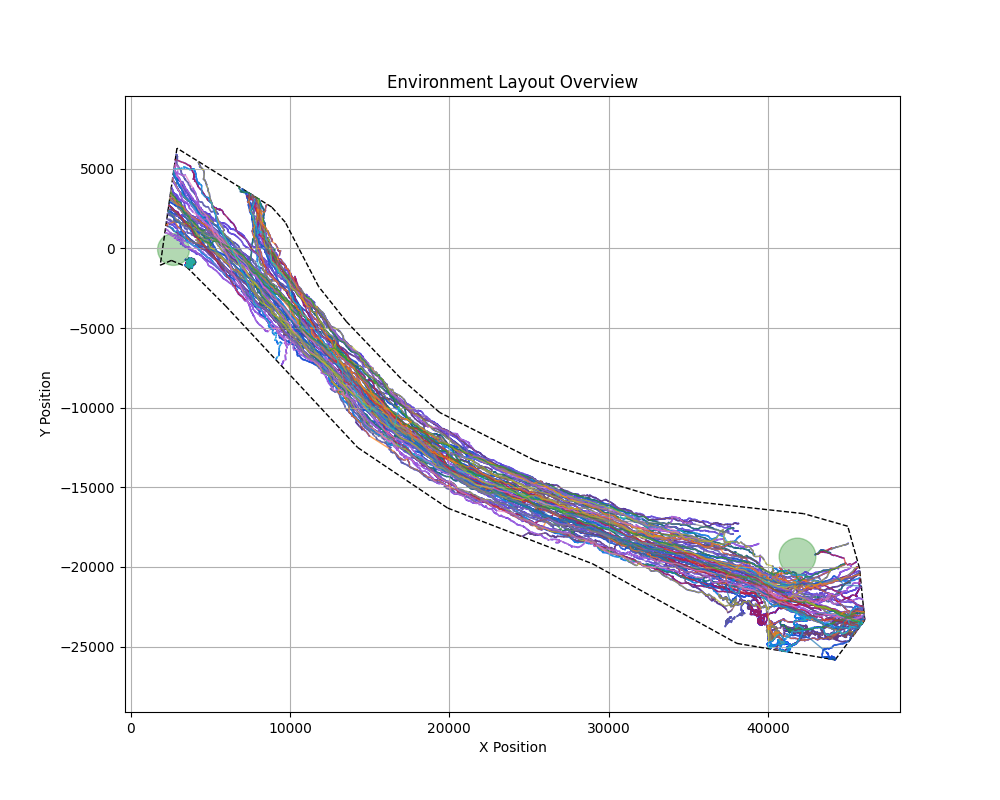}
\caption{\textbf{Environment layout overlaid with trajectories of 100 pedestrians, illustrating movement patterns and spatial interactions within the corridor region.}
\label{fig:environment}}
\end{figure}

As illustrated in Figure~\ref{fig:environment} data points located outside the defined environmental boundaries were excluded.  

Pillars are presented as green circles, while the surrounding spots such as shops, stairs, slopes, and exits are depicted as line segments.

\begin{table}[H]
    \centering
    \resizebox{\linewidth}{!}{
    \begin{tabular}{@{}llp{0.65\linewidth}@{}} % Adjust column widths as needed
    \toprule
    \textbf{Item} & \textbf{Value}     & \textbf{Short Description} \\ 
    \midrule
    Environment Size & ~5m $\times$ 55m & A corridor environment measuring approximately 5 meters in width and 55 meters in length. \\ 
    Number of Pillars & 2           & Number of structural pillars located within the environment. \\
    Total Number of Agents & 1,620     & Total number of unique agents observed in the dataset. \\
    Min Records per Agent & 32           & Minimum number of recorded data points for a single agent. \\
    Max Records per Agent & 91,271           & Maximum number of recorded data points for a single agent. \\
    Average Records per Agent &  1156.57         & Mean number of data records collected per agent across the dataset. \\
    Recording Frequency &  0.102 seconds & Average time interval between consecutive data recordings for each agent. \\
    Total Covering Duration & 1.05 hours & Total time span covered by the dataset across all agents. \\
    Total Number of Records & 1,873,638     & Total number of trajectory data points recorded in the dataset. \\
    Number of Unique Spots & 21          & Number of distinct spatial reference points (spots) defined by line segments within the environment. \\
    \bottomrule
    \end{tabular}
    }
    \caption{Summary of pedestrian trajectory statistics on May 8, 2013, during the 12:00–13:00 time interval.}
    \label{tab:dataset_stats}
\end{table}

Table~\ref{tab:dataset_stats} summarizes key statistics of the pedestrian trajectory dataset used for the study.

The group files contain annotations for the groups present on a given day. Only pedestrians in groups are listed, while those walking alone are excluded. Each row represents a single tracked pedestrian within a group and includes the following fields (space-separated values): $PEDESTRIAN-ID$, \textit{$GROUP-SIZE$}, \textit{$PARTNER-ID-1$}... (list of IDs of all other pedestrians in the group) as described in \url{https://dil.atr.jp/sets/groups/}

We use these group annotations to validate the detected group formations and to distinguish single pedestrians from group members.

\subsection{Scene Preparation for Space Utilization Analysis}

To analyze space utilization and extract emergent pedestrian behaviors, we implemented a structured analysis pipeline based on the flock detection framework proposed in \cite{amartaivan2025flock}.

\subsubsection{Pair-Based Group Classification Procedure}

The detection of group pedestrians—i.e., the classification of agents as singles or group members—requires the following preprocessing and model inference steps:

\begin{itemize}
    \item \textbf{Time Binning:} Pedestrians are grouped into fixed-size time interval bins (\texttt{TIME\_INTERVAL}) based on their starting timestamps.
    
    \item \textbf{Sequence Filtering:} Only pedestrians with trajectory sequences longer than a predefined threshold (\texttt{MIN\_SEQUENCE\_LENGTH}) are retained for further analysis.
    
    \item \textbf{Pair Prediction:} A pairwise classification model (Transformer-based in this study) is used to evaluate whether two agents belong to the same group. We adopt the feature design proposed in \cite{amartaivan2025flock} to train and apply this model.
    
    \item \textbf{Confidence Thresholding:} Predicted pairwise group relationships are filtered using a confidence score threshold. In this study, a threshold of 0.9 was applied to ensure high-quality pair predictions.
    
    \item \textbf{Flock Detection:} Based on confident pairwise links, pedestrians are clustered into groups (flocks) within each time bin using a union-find procedure.
\end{itemize}

The process begins by grouping agents into fixed-size time intervals (bins) according to their starting timestamps. Trajectories are filtered based on a minimum sequence length criterion to ensure sufficient temporal information.

Pairwise relationships between agents are classified using a pre-trained Transformer-based model trained on manually labeled pair data. Detected pairs are aggregated to form flock candidates, which are further refined to identify cohesive groups. The resulting group and single classifications are then used to conduct spatial and behavioral interaction analyses across time bins.

The details of these steps are described below.

\subsubsection*{Time Bin Preparation}

To enable consistent analysis and group detection, the trajectory data was segmented into fixed-length temporal windows, referred to as \textit{time bins}. Each bin represents a one-minute interval, and all agents whose trajectories begin within the same bin are grouped together for further processing. This segmentation is critical for both scalability and real-time applicability, as it allows localized, parallel analysis of group formation and interaction dynamics over time.

In addition to time binning, a sequence length threshold was applied to ensure that only trajectories with sufficient temporal continuity are included in downstream analysis. Sequence lengths of 60, 100, and 200 frames (equivalent to 20, 33, and 67 seconds at 3 Hz sampling rate) were tested to compare how different temporal resolutions affect the group classification quality and behavioral analysis.
% ----------------- Table 1 -------------------
\begin{table}[H]
    \centering
    \begin{tabular}{@{}ccccc@{}} 
        \toprule
        \textbf{Sequence Length} & \textbf{Total \# of Agents} & 
        \textbf{Min Agents per Bin} & \textbf{Max Agents per Bin} & \textbf{Mean Agents per Bin} \\
        \midrule
        60  & 1534 & 4  & 50 & 24.34 \\
        100 & 1491 & 3  & 50 & 23.67 \\
        200 & 1400 & 3  & 46 & 22.22 \\
        \bottomrule
    \end{tabular}  
    \caption{Agent statistics per sequence length across 63 time bins.}
    \label{tab:bin_vs_time}
\end{table}

Table~\ref{tab:bin_vs_time} summarizes the resulting statistics for each sequence length across 63 one-minute time bins. As expected, increasing the sequence length threshold reduces the total number of eligible agents due to the filtering of short or incomplete trajectories. While this trade-off reduces sample size, it improves the temporal stability of agent interactions and the reliability of extracted metrics, especially for behavioral indicators such as velocity change or heading deviation which require sufficient observation windows.

The mean number of agents per bin remains relatively stable across thresholds, indicating that longer sequences do not drastically reduce bin-level diversity. However, the minimum number of agents per bin can drop as low as 3–4, which may affect group detection performance in sparsely populated intervals. In future work, dynamic bin sizing or adaptive sequence filtering may be explored to mitigate these limitations.

This binning and filtering stage serves as the foundation for subsequent stages including pair classification, flock detection, and metric computation.

% ----------------- Figure -------------------
\begin{figure}[H]
\centering
\includegraphics[width=0.7\linewidth]{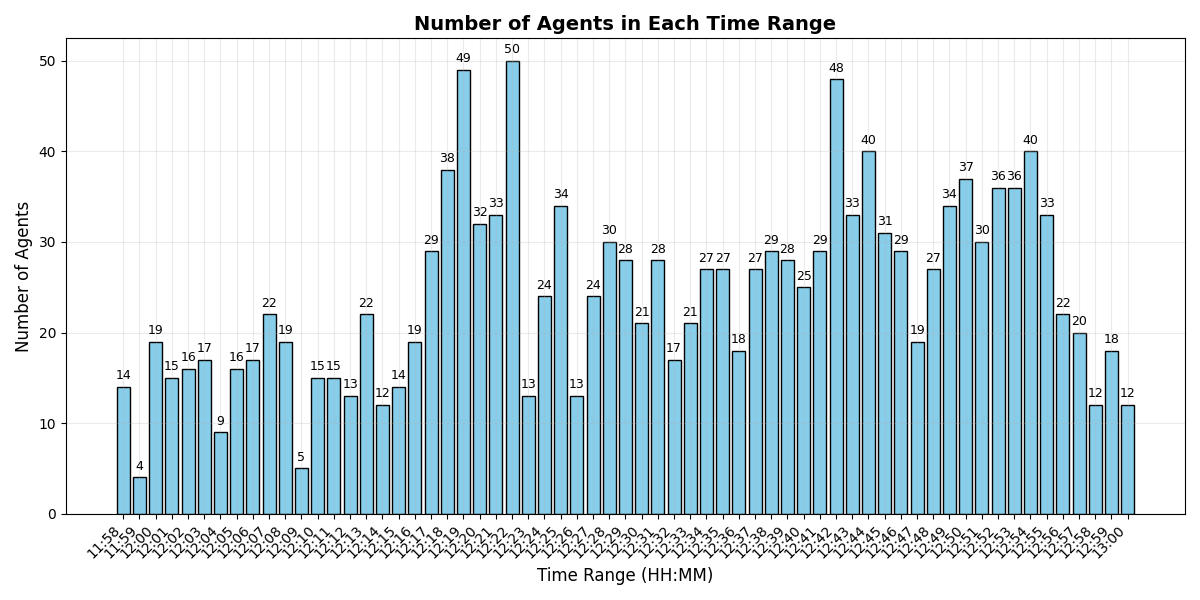}
\caption{\textbf{Sample bin visualization} with a time interval of 1 minute and sequence length of 60. Some agents are excluded based on the minimum sequence length criterion.}
\label{fig:sample_bin}
\end{figure}

\subsubsection*{Binary Classification Model Training}

To identify flock membership between pairs of pedestrians, we train a binary classification model based on the Transformer architecture. The input feature set is adapted from \cite{amartaivan2025flock}, and includes the following pairwise descriptors:

\begin{itemize}
    \item \textbf{Inter-distance:} Average Euclidean distance between two agents over a given sequence.
    \item \textbf{Time difference:} Offset between the starting timestamps of each agent.
    \item \textbf{Velocity difference:} Difference in average speed between the two trajectories.
    \item \textbf{Motion angle difference:} Difference in motion direction based on trajectory vectors.
    \item \textbf{Facing angle difference:} Difference in the orientation (facing) of each agent.
    \item \textbf{DTW distance:} Dynamic Time Warping similarity measure between the trajectories.
\end{itemize}

The dataset is constructed by selecting valid pedestrian pairs with sufficient sequence length and assigning a binary label: pairs identified as flocks are labeled \texttt{1}, while randomly sampled non-pair pedestrian combinations are labeled \texttt{0}. A 50:50 balanced dataset is maintained to avoid bias during training.

Model training is performed separately for three different sequence lengths: 60, 100, and 200 frames. For each configuration, we run 20 trials with varying hyperparameters (e.g., batch size, hidden layer size). The learning rate is fixed at 0.001, and early stopping is applied to prevent overfitting during training (maximum 1000 epochs). The best-performing configuration per sequence length is reported below.

% ----------------- Table 2 -------------------
\begin{table}[H]
    \centering
    \begin{tabular}{@{}ccccc@{}} 
        \toprule
        \textbf{Sequence Length} & \textbf{Total Samples} & 
        \textbf{Training Samples} & \textbf{Test Samples} & \textbf{Training Accuracy} \\
        \midrule
        60  & 3541 & 2833 & 708 & 92.701\% \\
        100 & 3453 & 2762 & 691 & 92.75\% \\
        200 & 3168 & 2534 & 634 & 93.56\% \\
        \bottomrule
    \end{tabular}  
    \caption{Training statistics and accuracy for pair classification using different sequence lengths.}
    \label{tab:seq_vs_sample}
\end{table}

As shown in Table~\ref{tab:seq_vs_sample}, increasing the sequence length leads to slightly higher classification accuracy. This trend suggests that longer sequences improve temporal consistency, which in turn enhances the reliability of pair-based relationship inference.

\subsubsection*{Flock Detection Results by Bins}

\begin{table}[H]
    \centering
    \begin{tabular}{@{}cccc@{}} 
        \toprule
        \textbf{Sequence Length} & \textbf{Total Agents} & 
        \textbf{Flock-Classified Agents (\%)} & \textbf{Detection Runtime (min)} \\
        \midrule
        60  & 1534 & 241 (15.71\%) & 50.22 \\
        100 & 1490 & 290 (19.46\%) &  120.21\\
        200 & 1402 & 261 (18.61\%) & 220.23\\
        \bottomrule
    \end{tabular}  
    \caption{Flock detection summary across different sequence lengths. Runtime indicates the wall-clock time for the detection process.}
    \label{tab:flock_detection}
\end{table}

As shown in Table~\ref{tab:flock_detection}, flock detection performance varies with the sequence length used for analysis. The \textit{Sequence Length} denotes the number of consecutive frames forming an observation window, where longer sequences capture more extended pedestrian dynamics but require more data. The \textit{Total Agents} decreases slightly as the sequence length increases, since longer windows filter out pedestrians with shorter trajectories. The \textit{Flock-Classified Agents (\%)} indicates the proportion of agents identified as belonging to flocks. Interestingly, increasing the sequence length from 60 to 100 frames raises the flock classification rate (from 15.71\% to 19.46\%), suggesting that additional temporal context improves group detection. However, at 200 frames the rate slightly drops (18.61\%), likely because overly long windows dilute local interaction patterns and exclude short trajectories. Finally, the \textit{Detection Runtime (min)} highlights the computational trade-off: runtime grows nonlinearly with sequence length (from 50.22 minutes at 60 frames to 220.23 minutes at 200 frames), emphasizing the balance between richer temporal context and computational efficiency.

\textbf{Sample Bin Visualization of Classified Trajectories: Group vs. Single Pedestrians
}

\begin{figure}[H]
\centering
\includegraphics[width=0.8\linewidth]{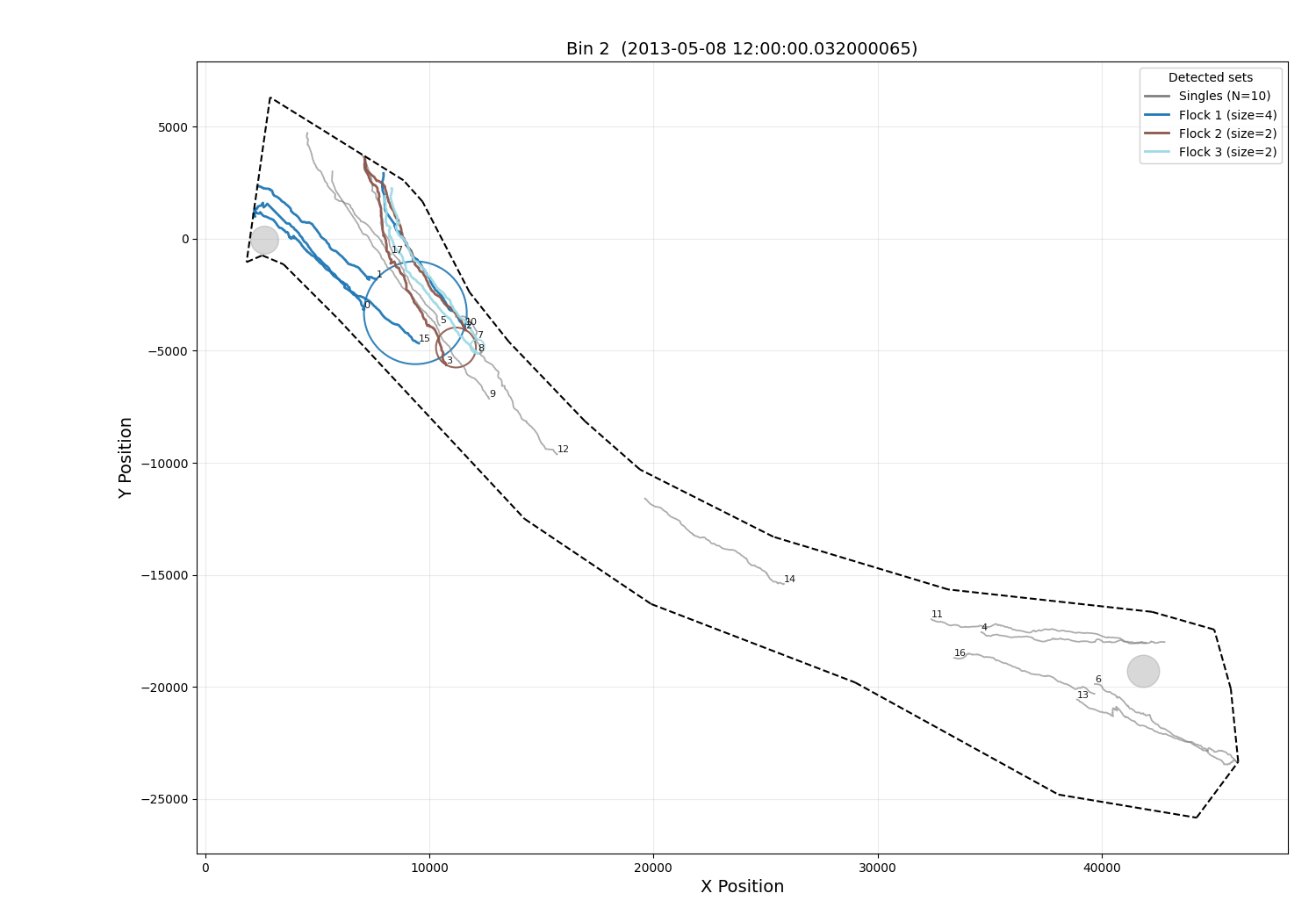}
\caption{\textbf{Sample classification results of pedestrian trajectories using the flock detection method. The visualization shows a total of 18 pedestrians, including 8 identified group members and 10 single pedestrians.}\label{fig:sample_flocks}}
\end{figure}

\section{Analysis Methods \& Preliminary Results}

To investigate how group and single pedestrians differ in their interactions with the environment and with each other, we designed a structured analysis pipeline consisting of two main components: spatial utilization metrics and behavioral pattern extraction. The pipeline is applied per time bin across multiple trajectory sequence lengths (60, 100, and 200 frames) to assess the impact of temporal granularity.

\subsection{Spatial Utilization Metrics}

Spatial usage is quantified based on the occupied area and density of pedestrian trajectories. The following metrics are computed per group or individual:

\begin{itemize}
    \item \textbf{Convex Hull Area:} Total area covered by a group or pedestrian trajectory over a sequence window.
    \item \textbf{Smallest Enclosing Circle (SEC):} The minimum radius circle enclosing all trajectory points, used to approximate spatial footprint.
    \item \textbf{Heatmap Density:} Aggregated position heatmaps over the environment to visualize space utilization over time.
\end{itemize}

These metrics offer insight into how different pedestrian types distribute themselves spatially, and whether group formations introduce crowding or spatial dominance effects. Spatial summaries are visualized using overlays on the environmental layout.

\begin{figure}[H]
    \centering
    \includegraphics[width=0.75\linewidth]{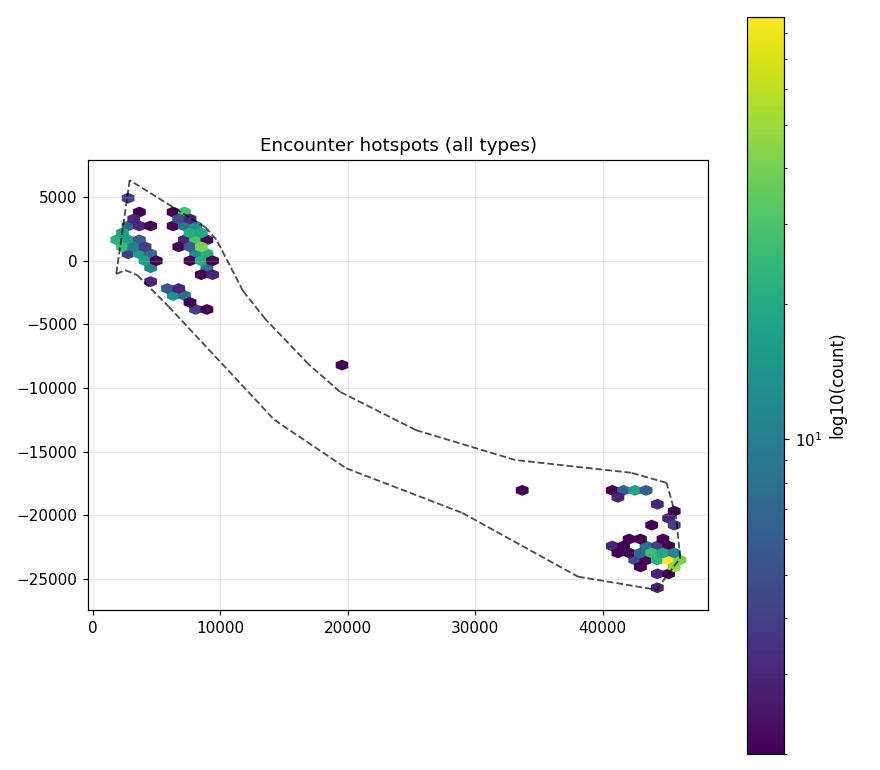}
    \caption{
        \textbf{Spatial encounter hotspots for all interaction types (sequence length = 60).}
        Hexbin plot showing spatial distribution of interaction events across the environment. Color represents the logarithm of event density (log$_{10}$ count).
        Dashed lines denote the spatial boundary of the environment. Dense interaction zones are found near structural bottlenecks and transition points.
    }
    \label{fig:encounter_hotspots}
\end{figure}

\begin{figure}[H]
    \centering
    \includegraphics[width=\linewidth]{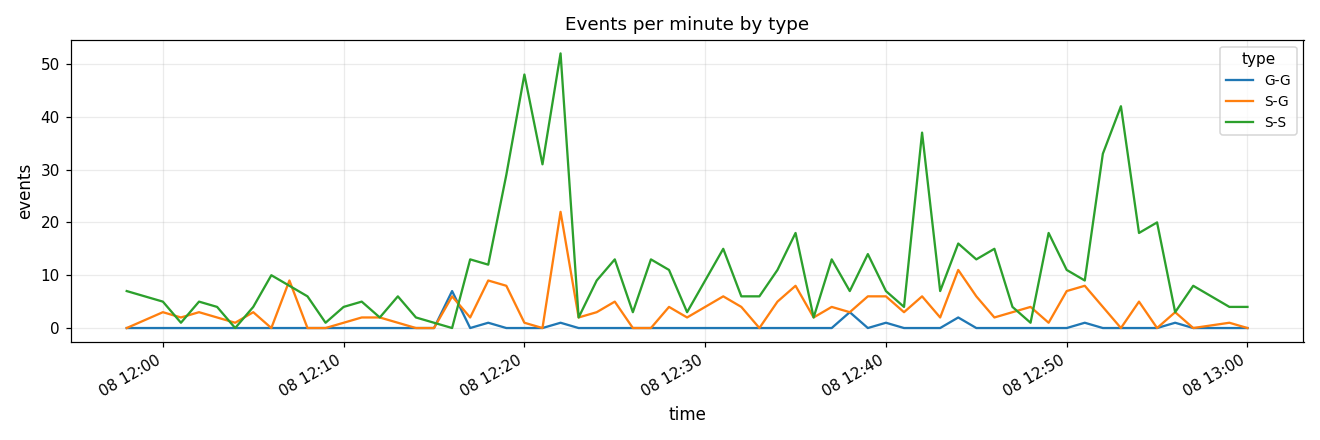}
    \caption{
        \textbf{Temporal distribution of interaction events (sequence length = 60).}
        Number of interaction events per minute across the 1-hour observation window (12:00–13:00) for each interaction type: G-G, S-G, and S-S. 
        S-S interactions occur most frequently, followed by S-G. G-G encounters are relatively rare but steady across time.
    }
    \label{fig:interaction_timeline}
\end{figure}

\subsection{Emergent Behavioral Pattern Extraction}

To evaluate interaction behavior and movement adaptations, we extract the following metrics for each pedestrian and pedestrian pair:

\begin{itemize}
    \item \textbf{Minimum Encounter Distance:} The closest point-to-point proximity during an interaction episode.
    \item \textbf{Speed Change ($\Delta v$):} Difference in average velocity before and after interaction with another agent or group.
    \item \textbf{Heading Change ($\Delta \theta$):} Variation in motion direction computed via consecutive heading angles.
    \item \textbf{Trajectory Straightness:} Ratio of the Euclidean distance between start and end points to the actual trajectory length.
    \item \textbf{Clearance Radius:} Distance maintained from nearby pedestrians, computed per timestamp.
\end{itemize}

Interaction types are categorized into:
\begin{itemize}
    \item \textbf{S–S:} Single-to-Single encounters
    \item \textbf{S–G:} Single-to-Group encounters
    \item \textbf{G–G:} Group-to-Group encounters
\end{itemize}

Each metric is computed per interaction and aggregated by interaction type and time bin. Per-event visualizations are generated for qualitative insights and outlier detection.

\begin{figure}[H]
    \centering
    \includegraphics[width=\linewidth]{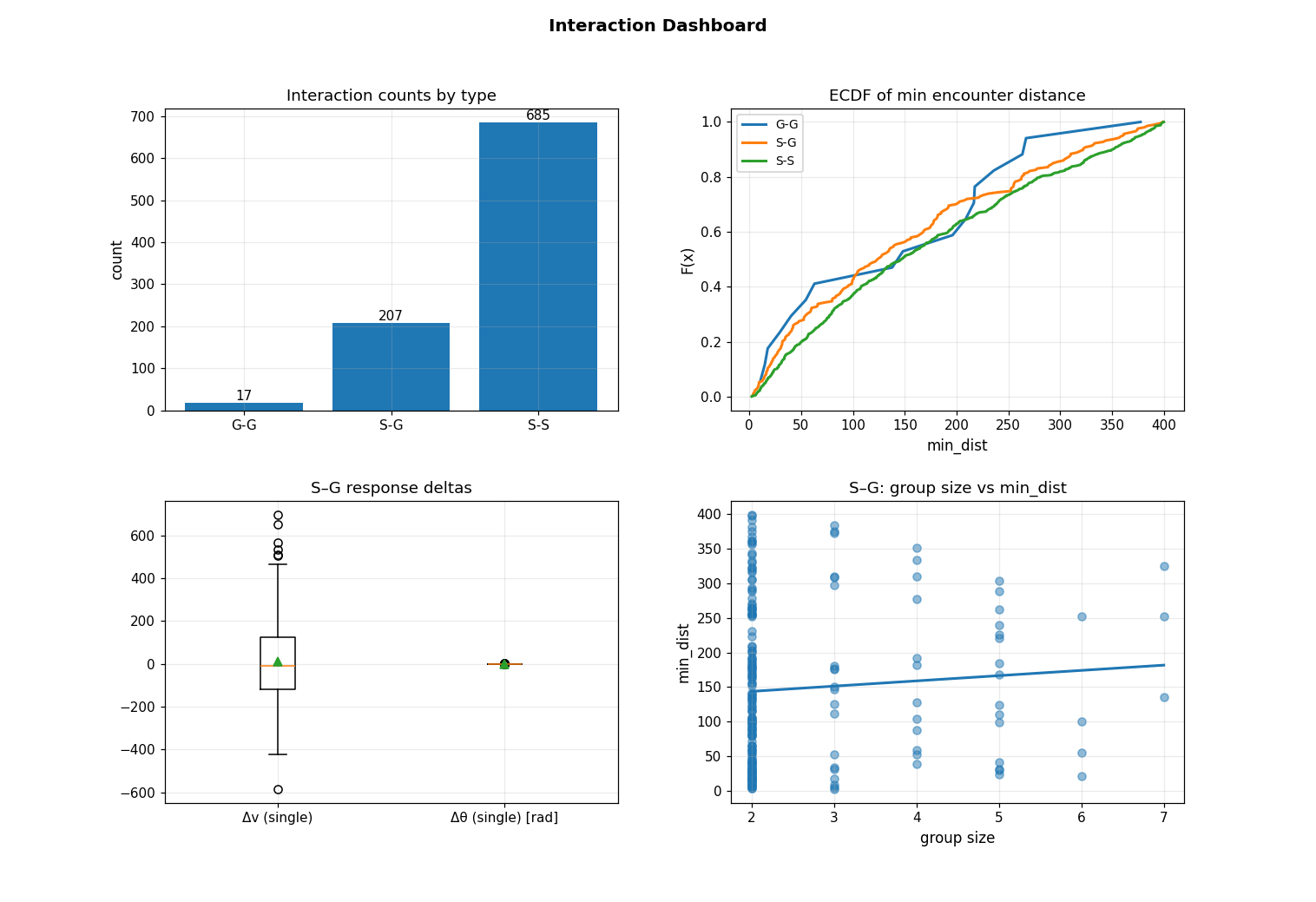}
    \caption{
        \textbf{Interaction Dashboard (sequence length = 60).} 
        Top-left: Histogram showing interaction event counts by type — Group-Group (G-G), Single-Group (S-G), and Single-Single (S-S). 
        Top-right: Empirical cumulative distribution function (ECDF) of minimum encounter distance for each interaction type. 
        Bottom-left: Box plots of $\Delta v$ (speed change) and $\Delta \theta$ (angular change in radians) for single pedestrians during S-G interactions, illustrating variability in response.
        Bottom-right: Scatter plot with regression line showing relationship between group size and minimum encounter distance in S-G interactions.
    }
    \label{fig:interaction_dashboard}
\end{figure}

\section{Discussion}

The preliminary findings underscore distinct spatial and behavioral differences between group and single pedestrians. Group pedestrians consistently exhibit larger spatial footprints, cohesive movement patterns, and a tendency to maintain formation even in spatially constrained environments. These group behaviors often influence the movement of surrounding single pedestrians, resulting in observable effects such as detours, yielding behavior, and reductions in speed. In contrast, single pedestrians demonstrate more adaptive pathfinding but are often forced to modify their trajectories in response to group presence.

Interaction typologies such as S--G (single-to-group) reveal asymmetries in agency and spatial negotiation, where single pedestrians often defer to group formations. G--G (group-to-group) interactions occasionally produce structured standoffs, with both groups attempting to preserve internal coherence despite spatial competition.

These behavioral distinctions hold valuable implications for both agent-based simulation and pedestrian-oriented space design. The extracted metrics—ranging from spatial occupancy to interaction-induced heading shifts—can inform the calibration of heterogeneous agents and local decision rules. From a design validation perspective, understanding how group pedestrians alter local flow dynamics can aid in setting corridor widths, optimizing bottleneck transitions, and planning for crowd resilience.

However, several limitations remain. The current approach relies on fixed thresholds (e.g., pair confidence score) and does not incorporate semantic role identification (e.g., friends, families, tour groups). Moreover, the analysis is constrained to a single environment. Future work will address these gaps by incorporating learning-based group role classifiers, expanding multi-environment datasets, and validating inferred behaviors through simulation benchmarks.

\section{Conclusion}

This paper presented a foundational study on the comparative analysis of space utilization and emergent behavior patterns between group and single pedestrians, using real-world trajectory data segmented by time bins. Leveraging a Transformer-based pair classification model and structured metric computation, we classified pedestrian groups and extracted key spatial and behavioral indicators.

Our contributions include: (i) a robust, sequence-based pipeline for real-time group classification, (ii) a unified framework of space utilization and behavioral metrics such as SEC area, $\Delta v$, $\Delta \theta$, and clearance radius, and (iii) a preliminary analysis highlighting the effects of group formations on pedestrian navigation and interaction.

While detailed metric evaluation and modeling applications are left for future versions, this work establishes the groundwork for scalable, behavior-aware pedestrian simulation and data-driven spatial design validation in public environments.

\bibliographystyle{unsrt}
\bibliography{references}

\end{document}